\documentclass[10pt,twocolumn,letterpaper]{article}

\usepackage{cvpr}              
\usepackage{balance} 
\usepackage{stfloats}
\usepackage{amsmath,amssymb,stmaryrd} 
\usepackage{graphicx}
\usepackage{multirow}
\usepackage{multicol}
\usepackage{makecell}
\usepackage{threeparttable}
\usepackage{caption}
\usepackage{stfloats}
\usepackage{cuted}
\usepackage{capt-of}
\usepackage{arydshln}

\usepackage{multirow}

\usepackage{multirow}

\definecolor{cvprblue}{rgb}{0.21,0.49,0.74}
\usepackage[pagebackref,breaklinks,colorlinks,allcolors=cvprblue]{hyperref}

\def\paperID{5112} 
\def\confName{CVPR}
\def\confYear{2026}

\title{NumeriKontrol: Adding Numeric Control to Diffusion Transformers for Instruction-based Image Editing}

\author{Zhenyu Xu\\
East China Normal University\\
\and
Xiaoqi Shen\\
South China University of Technology\\
\and
Haotian Nan\\
East China Normal University\\
\and
Xinyu Zhang\\
East China Normal University\\
}
\begin{document}


\maketitle

\begin{strip}\centering
    \includegraphics[width=\linewidth]{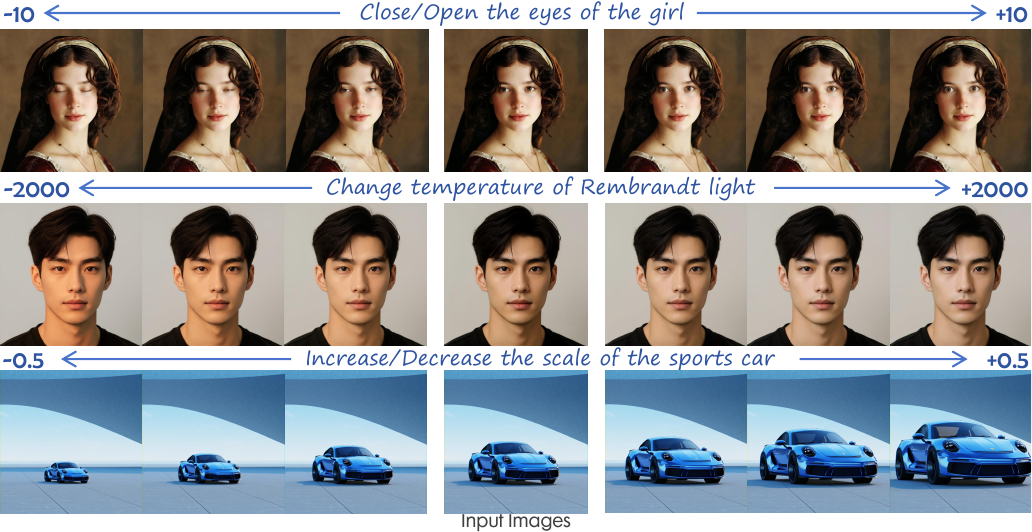}
    \vspace{-4mm}
    \captionof{figure}{
        NumeriKontrol generates precise editing trajectories across diverse attributes conditioned on a source image, editing instruction, and numeric strength. Trained on our synthesized dataset, the model enables precise control over low-level and high-level editing tasks.
    }
    \label{fig:teaser}
\end{strip}

\begin{abstract}

Instruction-based image editing enables intuitive manipulation through natural language commands. However, text instructions alone often lack the precision required for fine-grained control over edit intensity. We introduce NumeriKontrol, a framework that allows users to precisely adjust image attributes using continuous scalar values with common units. NumeriKontrol encodes numeric editing scales via an effective Numeric Adapter and injects them into diffusion models in a plug-and-play manner. Thanks to a task-separated design, our approach supports zero-shot multi-condition editing, allowing users to specify multiple instructions in any order. To provide high-quality supervision, we synthesize precise training data from reliable sources, including high-fidelity rendering engines and DSLR cameras. Our Common Attribute Transform (CAT) dataset covers diverse attribute manipulations with accurate ground-truth scales, enabling NumeriKontrol to function as a simple yet powerful interactive editing studio. Extensive experiments show that NumeriKontrol delivers accurate, continuous, and stable scale control across a wide range of attribute editing scenarios. These contributions advance instruction-based image editing by enabling precise, scalable, and user-controllable image manipulation.

\end{abstract}    
\section{Introduction}
\label{sec:intro}


Recent advancements in text-to-image generative models~\cite{DDPM, SD} have enabled instruction-driven image editing~\cite{IP2P}, allowing users to modify images via natural language instructions. However, effective editing requires controlling not only \emph{what} to edit but also \emph{how much}. For instance, designers may need to precisely adjust a character's smile, while photographers may require fine-grained control over the light effect of the image~(Fig.~\ref{fig:teaser}). This numeric precision in image editing remains largely underexplored, yet is crucial for bridging intuitive language interfaces with fine-grained parametric control.


Previous researches achieving fine-grained control via editing instructions mainly focus on continuous editing. Early GAN-based approaches~\cite{styleflow, ganspace, styleclip, precisecontrol, stylegan24} addressed the challenge through latent space manipulation for continuous attribute control. The effectiveness of these methods highly depend on the smoothness of latent space. Moreover, each GAN model should train from scratch. Other solutions based on diffusion models incorporating strength parameters~\cite{marble, alchemist} have demonstrated effective material editing, but they only limit on specific domain. Kontinuous Kontext~\cite{kontinuous} extends this paradigm to a unified instruction-based editing framework across diverse scenarios. It takes advantage of the modulation space introduced in Diffusion Transformers~(DiT). However, the reliance of Kontinuous Kontext on morphing-synthesized training data compromises numeric fidelity. Meanwhile, all of these methods edit the image with a normalized parameter ranging from $0$ to $1$. This ambiguates the actual effect of the instruction. More accurate instruction-based method is needed for better numeric controlled editing. 

Despite the demand for the fine-grained editing, several obstacles are essential to step over. The basic requirement is to preserve the image consistency while only attributes in the image changes along the input number. Additionally, the nonlinearity of the text embedding space is a major problem, limiting the effectiveness of introducing numbers into the prompt directly. Simultaneously, the unit in various scenarios differs from each other. The same numbers from two dimensions presents absolutely differed visual effects on the same image. Of particular concern is that there is no previous dataset indexed with accurate attribute transitions. Previous approaches~\cite{IP2P, kontinuous} synthesize the dataset through Large Language Models~(LLM), or provide a domain specific dataset~\cite{marble, alchemist}.


In this paper, we present NumeriKontrol, an instruction-based editing model that enables explicit quantitative control through parametric specifications in text prompts with diverse units. Utilizing the editing priors that ensure object and character preservation of FLUX.1 Kontext~\cite{kontext}, we introduce an effective numeric adapter. It encodes parametric values from prompts into embeddings, which are fused with task IDs for isolating different units. These numeric embeddings are as the same dimension as embeddings of instructions. The embeddings are injected into the DiT models by concatenating it to the text tokens. Benefit from such an In Context Learning~(ICL) approach, NumeriKontrol support zero-shot multi-numeric instruction editing. As the units are separated, the model produces identical results however the order the instructions are. To train NumeriKontrol, we construct the Common Attribute Transformation (CAT) dataset, comprising fundamental editing scenarios with ground-truth numeric value annotated. Crucially, all edited images are sourced from rendering engines or camera captures rather than synthetic interpolations for LLM-generated, ensuring the model learns genuine correspondences between numerical parameters and their visual manifestations. Meanwhile, the whole image maintains consistency with only the attribute mentioned in the instruction modifying.




Our main contributions can be summarized as follows:
\begin{itemize}
    \item We introduce the task of precise numeric control in instruction-based image editing while maintaining high consistency across the entire image.
    \item We propose a plug-and-play adapter for injecting numeric information into existing DiT models, enabling precise numeric control via an in-context learning paradigm.
    \item We synthesize a comprehensive dataset for common editing scenarios, covering general image editing operations with ground-truth control annotations. Extensive experiments across diverse editing tasks demonstrate that NumeriKontrol achieves superior precision and consistency compared to existing methods.

\end{itemize}

\section{Related Works}
\label{sec:related-works}

\subsection{Diffusion Models} 

Diffusion models have emerged as a prominent class of generative models, evolving from their foundational denoising probabilistic frameworks to state-of-the-art architectures~\cite{DDPM, ddim, SD}. These models have demonstrated remarkable efficacy across diverse image generation tasks, including text-to-image synthesis~\cite{sdxl, photorealistic} and image editing~\cite{IP2P, masactrl, imagic}. Through iterative noise refinement processes, diffusion models achieve unprecedented fidelity and diversity in generated content. Early latent diffusion models (LDMs) leveraged CLIP~\cite{clip} to align text and image representations within a shared latent space. Recently, a paradigm shift has occurred in diffusion model architectures, transitioning from conventional U-Net-based designs to transformer-based frameworks. This architectural evolution, exemplified by models such as SD3~\cite{rf} and FLUX~\cite{flux2024}, adopts Diffusion Transformer~\cite{dit} structures that enable enhanced image quality, higher-resolution synthesis, and improved processing of complex conditional inputs. These transformer-based models modulate text tokens extracted from CLIP encoders with embeddings from languange models and noised latent representations.

\begin{figure*}[ht]
    \centering
    \includegraphics[width=\linewidth]{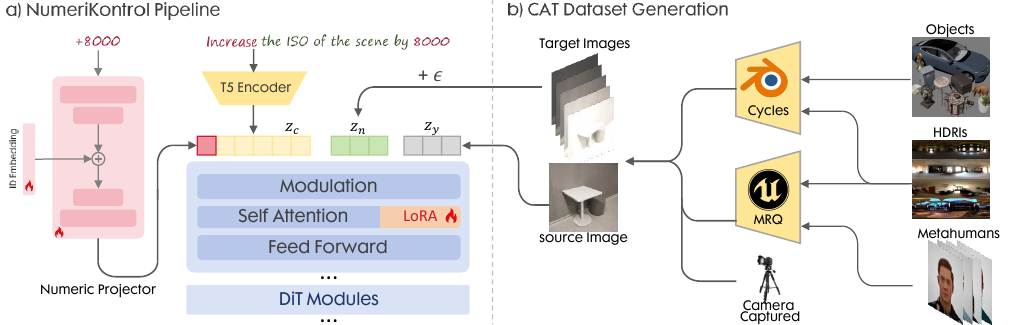}
    \vspace{-4mm}
    \captionof{figure}
    {
    \textbf{a) Overall pipeline of NumeriKontrol.} Numeric information is extracted from the editing instruction and encoded, subsequently being fused with the task ID embedding. Positive and negative numeric editing operations share an identical task ID. We construct a comprehensive operation dataset comprising both synthetically generated and real-captured images. \textbf{b) Dataset generation.} Dataset diversity is ensured through heterogeneous data sources. During training, origin and context images are randomly sampled from the attribute animation sequences.
    }
    \label{fig:pipeline}
\end{figure*}

\subsection{Controllable Generation} 

The synthesis process of diffusion models can be controlled across multiple dimensions. LoRA~\cite{lora} enables fine-tuning of LDMs to generate images with desired characteristics. ControlNet~\cite{controlnet} and subsequent works ~\cite{ma2024followyouremoji, ma2025controllable, ma2025followcreation, ma2025followfaster, ma2025followyourclick, ma2025followyourmotion, song2024processpainter, zhang2024stable, zhang2025stable, chen2025transanimate}   leverages structural information by incorporating trainable copies of the diffusion model's encoding layers to guide generation. T2I-Adapter~\cite{t2i-adapter} enhances computational efficiency through lightweight adapter modules. IP-Adapter~\cite{ipadapter} introduces disentangled cross-attention through an additional
encoder to fit for non-aligned condition tasks. With the architectural transition to Diffusion Transformers~\cite{dit}, recent approaches such as EasyControl~\cite{easycontrol} and OmniControl~\cite{ominicontrol} have achieved image-conditioned generation within MM-DiT frameworks and inspired subsequent works ~\cite{song2025omniconsistency, gong2025relationadapter, jiang2025personalized, wang2025diffdecompose, song2025makeanything, song2025layertracer, guo2025any2anytryon,  lu2025easytext, shi2025wordcon, shi2024fonts}. However, existing research predominantly focuses on image-based guidance, leaving a gap in numerical control methods that would enable more precise and quantitative manipulation of generated content.

\subsection{Instruction-based Image Editing} 

Instruction-based image editing, pioneered by InstructPix2Pix~\cite{IP2P}, enables image manipulation through natural language instructions. To achieve this, the authors curated a large-scale synthetic dataset comprising image-edit pairs generated via Prompt2Prompt~\cite{prompt2prompt}, with corresponding editing instructions synthesized by large language models (LLMs). More recent approaches~\cite{qwen-image, omnigen2, seedream, huang2025photodoodle, ma2023magicstick} have developed unified large-scale models capable of performing both generation and editing tasks \cite{ma2024followpose, ma2022visual, ma2023magicstick}. These models demonstrate proficiency across diverse editing operations, including subject-driven personalization, compositional scene editing, and instruction-guided modifications. Despite their impressive versatility in general-purpose editing scenarios, these models lack mechanisms for numerical control over editing attributes, thereby limiting their utility for applications requiring precise, fine-grained adjustments.

\section{Preliminary}
\label{subsec:preliminary}

\noindent \textbf{Problem Formulation}. For instruction-based image editing, a model synthesizes an edited image following a reference image $c$ and a text editing prompt $y$. It aims at learning the conditional distribution $p(x \mid c, y)$. We define our task by extending general instruction-based image editing with a numeric strength. For any editing instruction $y$ containing an attribute value $n$, the model aims at approximating the conditional distribution $p(x \mid c, y, n)$. Approaching the distribution by diffusion model~\cite{rf, SD, sdxl}, the target is to obtain a network $\epsilon_\theta(x(t), y, c, n)$ that predicting a noise. The result $x$ can be obtained via several denoising steps.

\noindent \textbf{FLUX.1 Kontext}. FLUX~\cite{flux2024} is a Text-to-Image model build upon Diffusion Transformers. FLUX.1 Kontext~\cite{kontext} fine-tunes the FLUX model to fit in image editing tasks. It takes a source image and an edit instruction as input and outputs the edited result. In this pipeline, the image and text are encoded as tokens and processed through visual and textual attention streams. Kontext extends to editing tasks by encoding the source (context) image with the autoencoder. The context token is then concatenated to the noise tokens, which are jointly processed in the visual stream. As FLUX.1 Kontext achieves the network $\epsilon_\theta(x(t), y, c)$, the problem switches to how to inject $n$ into the model.

\section{Method}
\label{sec:method}

In this section, we propose a numeric projector to encode numeric strength and integrate it into the MM-DiT architecture~(Sec.~\ref{subsec:numeric-adapter} and ~\ref{subsec:conditioning}). For multi-condition scenarios, we introduce a decoupled attention masking strategy to prevent cross-condition interference~(Sec.~\ref{subsec:multi-condition}). Finally, we describe the dataset construction combining rendering engine-generated samples and camera-captured photographs (Sec.~\ref{subsec:dataset}).

\subsection{Unit Isolated Numeric Projector}
\label{subsec:numeric-adapter}

Given that FLUX.1 Kontext provides a prior for image consistency, our objective is to inject numeric edit strength information into the Diffusion Transformers. Since numeric information is embedded within the instruction, we investigate whether fine-tuning on data with captions containing numbers suffices. Through empirical study, we fine-tune the FLUX.1 Kontext model and observe that direct learning of numeric transitions proves infeasible. Fine-tuning on such datasets results in the model generating averaged representations according to the prompts. Notably, numeric information is almost entirely discarded during the training process.

To address the limitation, we propose a lightweight numeric projector that explicitly encode strength information. This adapter encodes any strength value $\bar{n}$ from instruction $y$ into a feature representation $z_{\bar{n}} \in \mathbb{R}^{1 \times d}$ via several MLP layers. The major concern lies in that two attributes to be edited may have different units but same values. Consider moving an object 10cm forward and rotating 10 degrees along the Z-axis. If the number is encoded into one token identically, the token undertakes representing two definitely different attribute. Consequently, the model may rotate the object by 10 degree while the user intends to move forward. We therefore design a two-stage process with task fusion. Initially, the strength value $n$ is encoded into hidden states $H_n$. A trainable task-specific identifier $\mathbf{T}$ is embedded and fused with $H_n$ to accommodate diverse editing tasks. Subsequently, the hidden states are decoded through the MLP, yielding $z_{\bar{n}}$. The complete encoding process can be formulated as:
\begin{equation}
z_{\bar{n}}=\mathcal{D}(\mathcal{E}(n)+\operatorname{emb}(\mathbf{T})),
\end{equation}
\noindent where $\mathcal{E}$ and $\mathcal{D}$ denote MLPs functioning as encoder and decoder, respectively. When several numbers with same value but different units are encountered, the task embeddings separate these tokens into their own dimensions. It effectively avoids task-switching during inference. 

\subsection{Numeric Adaptation into MM-DiT}
\label{subsec:conditioning}

Within the FLUX.1 Kontext architecture, the model processes three distinct types of input features: text tokens $z_t$, noise representations $z_n$, and context tokens $z_y$. These input features are initially projected into their respective query $\mathbf{Q}$, key $\mathbf{K}$, and value $\mathbf{V}$ representations through learnable projection matrices, as formulated below:

\begin{equation}
\begin{aligned}
& \mathbf{Q}_t = W_\mathbf{Q} z_t, \mathbf{K}_t = W_\mathbf{K} z_t,  \mathbf{V}_t= W_\mathbf{V} z_t, \\
& \mathbf{Q}_n = W_\mathbf{Q} z_{ny}, \mathbf{K}_n = W_\mathbf{K} z_{ny}, \mathbf{V}_n= W_\mathbf{V} z_{ny}, \\
\end{aligned}
\end{equation}

\noindent where $z_{ny} = [z_n, z_y]$ denotes the concatenation of noise and context tokens. Subsequently, these projected features are concatenated across modalities to form unified representations $[\mathbf{Q}_t, \mathbf{Q}_n]$, $[\mathbf{K}_t, \mathbf{K}_n]$, and $[\mathbf{V}_t, \mathbf{V}_n]$, which are then processed through a self-attention, depicted in Fig.~\ref{fig:pipeline}~(a).

To exploit the in-context learning capabilities of MM-DiT, we introduce a modified attention formulation. This formulation directly integrates numeric tokens $\bar{n}$ into the text token stream. Concretely, we concatenate numeric token representations with text tokens prior to projection:

\begin{small}
\begin{equation}
\label{eq:qkv}
\mathbf{Q}_t= W_\mathbf{Q} [z_{\bar{n}}, z_t], \\
\mathbf{K}_t = W_\mathbf{K} [z_{\bar{n}}, z_t],\\
\mathbf{V}_t= W_\mathbf{V} [z_{\bar{n}}, z_t].
\end{equation}
\end{small}

Consequently, the original text query representation $\mathbf{Q}_t$ in Eq.~\ref{eq:qkv} is replaced by the extended representation $[\mathbf{Q}_{\bar{n}}, \mathbf{Q}_t]$, enabling the model to jointly attend to both numeric and textual information. To enable the DiT further understand the concatenation of the numeric token and text tokens, all $W_\mathbf{Q}, W_\mathbf{K}$ and $W_\mathbf{V}$ utilizes Low-rank Adaptations~\cite{lora} during model training.

For model training, we construct a paired dataset comprising quadruplets of ($x$, $c$, $n$, $y$), representing source image, editing instruction, numeric editing parameter and target edited image correspondingly. Given a training sample $(x, y, c, n)$ and a randomly sampled diffusion timestep t, we optimize the model parameters using the standard flow matching objective:

\begin{equation}
\mathcal{L}_\theta=\mathbb{E}_{t \sim p(t), y, c, n}||v_\theta\left(x^t, y, c, n, t\right)-(\epsilon-x)||_2^2,
\end{equation}
\noindent where $x^t$ represents the interpolated latent representation between the clean latent $y_c$ and Gaussian noise $\epsilon \sim \mathcal{N}(0,1)$, computed as $x^t=(1-t) x +t \epsilon$. Here, $v_\theta$ denotes the transformers in FLUX.1 Kontext parameterized by $\theta$.

\subsection{Decoupled Multi-numeric Editing}
\label{subsec:multi-condition}

\begin{figure}[t]
    \centering
    \includegraphics[width=\linewidth]{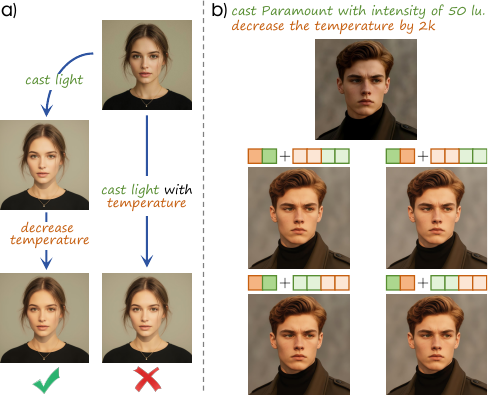}
    \captionof{figure}{\textbf{a)} Example of resolving coupled instruction. Training directly on such instructions fails, as the real start point of the edit is the image casted by a light but no the source image. \textbf{b)} Visualization of decoupled and order-independent editing through multiple numeric instructions.}
    \label{fig:multi-condition}
\end{figure}

Generally, some editing tasks cannot achieve by one attribute transformation, but can be decomposed into several simple tasks. For an instruction $y$, it may be actually separated into $y_0, y_1, ...,y_n$ to precisely describe the editing. As is shown in Fig.~\ref{fig:multi-condition}~(a), the instruction \textit{Cast a light with temperature} should be resolved as \textit{Cast a light with intensity of $||\bar{n}_0||$ lumen} and \textit{Change the temperature of the light by  $||\bar{n}_1||$ Kelvin}.Composed instructions make dataset construction and training more sophisticated, a multi-numeric editing is essential for such cases.

During training, each numeric token learns to interact with its corresponding task ID embedding, editing instruction template and context tokens. To enable multi-numeric conditioning, we develop a zero-shot multi-numeric paradigm inspired by EasyControl~\cite{easycontrol} in inference. For each sub-task, the numeric value is encoded as $z_{\bar{n}_i}, i \in N$ where $N$ represents number of tasks. All numeric tokens are concatenated as $[z_{\bar{n}_0}, z_{\bar{n}_1}, ... ,z_{\bar{n}_N}]$. Each task corresponds to an instruction template, the final instruction concatenated from all of the sub-instructions. 

In multi-numeric inference, cross-attention between numeric tokens from different conditions introduces undesired interference. A structured attention mask is employed that isolates different numeric tokens while allowing noise tokens to aggregate information from all sources. For $N$ numeric conditions, the self-attention sequence is constructed as:

\begin{equation}
\mathbf{Z} = [\mathbf{z}_{\bar{n}_1},\mathbf{z}_{\bar{n}_2},...,\mathbf{z}_{\bar{n}_m}, \mathbf{z}_t, \mathbf{z}_n],
\end{equation}

\noindent where $\mathbf{z}_{\bar{n}_i}, \mathbf{z}_t$, and $\mathbf{z}_n$ represent tokens for the $i$-th numeric condition, text tokens, and noise tokens, respectively. We define an attention mask $M \in\{0,-\infty\}^{n \times n}$ as:

\begin{small}
\begin{equation}
M_{i j}= \begin{cases}0, & \text{if } (i, j \in \mathcal{I}_t \cup \mathcal{I}_n) \text{ or } (i = j). \\ -\infty, & \text{otherwise}.\end{cases}
\end{equation}
\end{small}

\noindent where $\mathcal{I}_t$ and $\mathcal{I}_n$ denote the index sets of text and noise tokens. This masking ensures that  numeric tokens from different conditions remain isolated while noise tokens attend to all conditions. It prevents mutual interference while preserving independent conditional control. During our experiment, we found that the order of the editing prompts and numeric tokens does not influence the result, illustrated in Fig.~\ref{fig:multi-condition}~(b). As the numeric tokens are fused by the task embeddings, it only interacts with corresponding text tokens. This phenomenon lies on the editing-consistency prior of FLUX Kontext. It tends to predict an origin image if the editing is not available or understood by the model.

\subsection{Synthesize Attribute Dataset}
\label{subsec:dataset}

A heuristic approach to achieve continuous editing control is to generate intermediate images by interpolating between the source and edited images. Existing diffusion-based morphing methods~\cite{diffmorpher, freemorph} interpolate in the latent space, assuming semantic smoothness. However, this assumption does not hold universally. Such methods are sensitive to outliers and frequently produce artifacts in intermediate results, including missing objects and blurred content (Fig.~\ref{fig:dataset_compare}). As depicted in Fig.~\ref{fig:pipeline}~(b), we construct a dataset from two complementary sources to enable broader numeric control:

\begin{figure}[b]
  \centering
   \includegraphics[width=\linewidth]{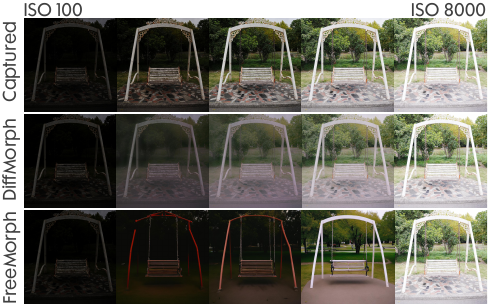}

   \caption{
   \textbf{Illustration of difference from physically synthesized data to morphing-generated intermediates.} Interpolation-based morphing methods fail to preserve visual fidelity even in straightforward low-level editing scenarios such as ISO modification. FreeMorph, a representative diffusion-based morphing technique, samples trajectories with substantial deviation from ground truth sequences.
   }
   \label{fig:dataset_compare}
\end{figure}

\noindent \textbf{Rendering Engine-Based Data Generation.} For editing operations difficult to capture in real-world settings, we refer to professional rendering engines, Blender and Unreal Engine, to generate physically plausible and visually realistic images. Cycles renderer and MRQ system with path tracing are used in these two engines respectively, ensuring high-quality render results. For object transformations,  Blender is chosen to manipulate 3D assets. We collect a large amount of FBX objects and HDRI backgrounds online. Each scene comprises an object (ranging from small items to vehicles) and one of HDRI backgrounds. For example, we apply rotations ranging from $-90$ degree to  $90$ degree, sampling 180 rotation states into dataset. For human face and pose animations, we employ the Unreal MetaHuman Editor to generate 100 distinct characters across various ages, each rendered in a different HDRI environments. Since scenarios as facial expressions and pose transition lack inherent numeric parameterization, we encode the origin status as 0 and map target expressions to ±10. The instruction of each scenario is human-designed for better understanding. For better generalization, the camera view point is selected randomly from the half-sphere centered at the object.
\begin{figure*}[t]
    \centering
    \includegraphics[width=\linewidth]{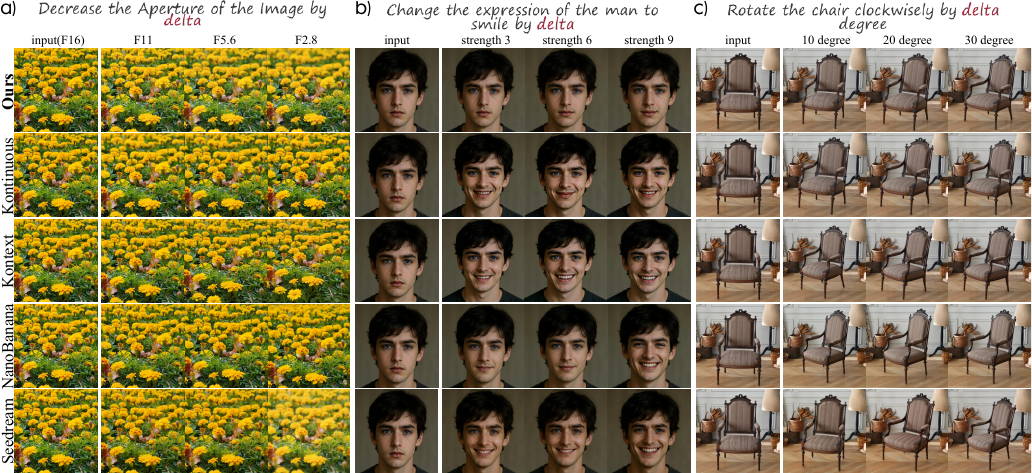}
    \captionof{figure}{\textbf{Visualization results on NumeriKontrol and baseline methods.} Various scenarios, including outdoor scene, portrait and object editing, are selected from the results. The \textit{"delta"} in each caption is replaced by the actual number above the images in NumeriKontrol.  Captions of other methods are tuned respectively. Compared to others, the smile of the man generated by NumeriKontrol is less prominent because the model learned from a dataset containing only subtle smile examples.
    }
    \label{fig:qualitative}
\end{figure*}

\noindent \textbf{Camera-Captured Dataset.} \noindent \textbf{Camera-Captured Dataset.} Rendering engines excel at geometric transformations but fail to reproduce realistic camera-specific effects. To enable our model to function as a versatile camera studio, we capture real photographic data using a Digital Single-Lens Reflex (DSLR) camera. This provides authentic samples for camera parameter adjustments. For instance, we collect 200 scenes with ISO values ranging from 100 to 8000 covering objects, indoor environments, and outdoor landscapes. Camera parameters typically exhibit nonlinear characteristics. We formulate the editing instruction as "Increase/Decrease the $\mathbf{P}$ by $\|\delta\|$", where $\mathbf{P}$ denotes the target parameter and $\delta$ represents the measured difference between source and target values. Other attributes are all captured in reasonable range. During training, the input and target image is randomly chosen and the $\delta$ is calculated online for better generalization.
\section{Experiments}

\subsection{Experiment Set Up}

\noindent \textbf{Implementation Details.} We employed official FLUX.1 Kontext~\cite{kontext} pretrained model as the base model. The Numeric Projector is a four layer MLP with dimensions $256 \rightarrow 256 \rightarrow 256 \rightarrow 4096$ in our experiments. The LoRA applied on attention layers of DiT is set to 128. During training, two A800 GPUs(80GB) are devoted into optimization progress, with a batch size of 4 per GPU, and a learning rate of 5e-5, training over 160,000 steps. During inference, the flow-matching sampling is applied with 28 sampling steps. 

\noindent \textbf{Baseline Methods.} We validate our method against several state-of-the-art methods. We choose FLUX.1 Kontext~\cite{kontext}, Kontinuous Kontext~\cite{kontinuous}, Nano Banana and Seedream 4.0~\cite{seedream} as our baseline. The classifier-free guidance~\cite{cfg} of FLUX.1 Kontext is set to 1.0. For Nano Banana and Seedream, official API and website is used in evaluation. As Kontinuous Kontext is not open-sourced, we implement it as is described in its paper. The models for Kontinuous are trained from a simpler CAT dataset, which removes the strength information from prompt and normalizes the parameters from $0$ to $1$. 

\noindent \textbf{Benchmarks.} We construct a test set for the CAT dataset, comprising representative examples across fundamental editing scenarios. Based on task complexity, two evaluation categories are established. The first category encompasses low-level editing tasks that applied globally to the image, specifically adjustments as ISO sensitivity, aperture settings, white  balance parameters and so on. The second category consists of high-level editing tasks that require spatial and semantic understanding, including facial expression manipulation, pose modification and other object-level operations. For diversity, an aspect of the dataset enable bidirectional manipulation as examples shown in Fig.~\ref{fig:teaser}. At last, totally 200 editing cases are included in the test set.

\noindent \textbf{Metrics.} To evaluate the quality and accuracy of results, we test the results on PSNR, SSIM~\cite{ssim}, LPIPS~\cite{lpips} and Pixelwise Mean Square Loss~(MSE). All metrics focus on the similarity and consistency against the ground truth.

\begin{table*}[htbp]
\centering
\caption{
\textbf{Quantitative Comparison.} The best of each metric is \textbf{emphasized}.
}
\begin{tabular}{llcccclcccc}
\toprule
\multirow{2}{*}{Method} &  & \multicolumn{4}{c}{Low Level Tasks} &  & \multicolumn{4}{c}{High Level Tasks} \\ 
    \cline{3-6} \cline{8-11}  &  & \multicolumn{1}{c}{PSNR~$\uparrow$} & \multicolumn{1}{c}{SSIM~$\uparrow$} & \multicolumn{1}{c}{MSE~$\downarrow$} & \multicolumn{1}{c}{LPIPS~$\downarrow$} & & \multicolumn{1}{c}{PSNR~$\uparrow$} & \multicolumn{1}{c}{SSIM~$\uparrow$} & \multicolumn{1}{c}{MSE~$\downarrow$} & \multicolumn{1}{c}{LPIPS~$\downarrow$} \\ 
    \hline \hline
    
FLUX.1 Kontext~\cite{kontext}          &  & 19.694  & 0.7169  & 0.0236 & 0.1264 &  & 18.214  & 0.7642   & 0.0289 & 0.1876 \\
Kontinuous Kontext~\cite{kontinuous}   &  & 21.452  & 0.7366  & 0.0225 & 0.1079  &  & 24.753 & 0.8624 & 0.0117 & 0.0755\\
Nano Banana                            &  & 18.272 & 0.5880 & 0.0413 & 0.2080  &  & 22.868  & 0.8688  & 0.0153 & 0.0984 \\
Seedream~\cite{seedream}               &  & 13.659  & 0.5337  & 0.0721 & 0.3048 &  & 17.873   & 0.8202  & 0.0279 & 0.1411  \\
\textbf{NumeriKontrol(Ours)}           &  & \textbf{22.678}  & \textbf{0.7633}  & \textbf{0.0190} & \textbf{0.0980}  &  & \textbf{27.560}   & \textbf{0.9120}   & \textbf{0.0084}  & \textbf{0.0541}    \\

\bottomrule
\end{tabular}
\label{tab:quantitative}
\end{table*}

\subsection{Qualitative Comparison}
\label{subsec:qualitative}

We conduct qualitative comparisons between NumeriKontrol and baseline methods including Kontext~\cite{kontext}, Kontinuous Kontext, Nano Banana and Seedream~\cite{seedream} to illustrate the effectiveness of NumeriKontrol. Three visual results covers different scenarios, including both low-level and high-level tasks. The results are shown in Fig.~\ref{fig:qualitative}. 

For aperture editing tasks shown in Fig.~\ref{fig:qualitative} (a), both LLM-based methods (Nano Banana and Seedream) fail to generate the desired effects, demonstrating inadequate understanding of the \textit{Aperture} concept. Alternatively, These methods  approximate similar results via similar instructions as \textit{blur the farther flowers.}, but the resulting images remain substantially suboptimal. FLUX.1 Kontext also lacks comprehension of aperture semantics; nevertheless, it preserves the original image rather than introducing erroneous modifications. In contrast, NumeriKontrol and Kontinuous Kontext demonstrate competency in aperture manipulation due to their training on CAT. However, our method achieves superior aperture-matching performance. A notable limitation of Kontinuous is need for strength normalization. It is incompatible with non-linear design of camera F-numbers. 

As illustrated in Fig.~\ref{fig:qualitative}~(b) and (c), all methods successfully execute edits in accordance with user prompts for high-level tasks. However, FLUX.1 Kontext fails to preserve numeric information from the instructions, generating identical images regardless of attribute modifications. When provided with prompts such as \textit{Change the expression of the man to smile by 0.3 strength. Full strength is 1.0}, Nano Banana and Seedream produce varied editing outcomes; the visual manifestations in the edited images do not exhibit a monotonic correspondence with user-specified strength values. Despite Kontinuous being trained on the same CAT dataset, its results deviate substantially from the ground truth.

\subsection{Quantitative Comparison}
\label{subsec:quantitative}

We present quantitative comparisons with baseline methods in Table.~\ref{tab:quantitative} for further validating the accuracy of our approach. Low-level tasks operate primarily in the pixel domain, applying global adjustments without altering image content, whereas high-level tasks modify localized regions and introduce semantic-level changes. Given the fundamental distinctions between these task categories, metrics are computed separately for each.

\begin{table}[t]
    \centering
    \caption{\textbf{Quantitative results of the ablative study}.The best of each metric is \textbf{emphasized}. Line 1-4 sequentially decreases the LoRA while line 5-7 sequentially removes the module from model. Two series of experiments are evaluated separately.
    }
    \resizebox{\linewidth}{!}{
    \begin{tabular}{lcccc}
    \toprule
    Methods & PSNR~$\uparrow$ & SSIM~$\uparrow$ & MSE~$\downarrow$ & LPIPS~$\downarrow$ \\
    \midrule
    \textbf{NumeriKontrol} & \textbf{25.119} & \textbf{0.8377} & \textbf{0.0137} & \textbf{0.0760}  \\
    \hdashline
    LoRA Rank 64 & 22.766 & 0.8004 & 0.0182 & 0.0898 \\
    LoRA Rank 32 & 21.080 & 0.7903 & 0.0206 & 0.1020 \\
    LoRA Rank 4 &  20.263 & 0.7691 & 0.0235 & 0.1399 \\
    LoRA Rank 0 &  19.671 & 0.7524 & 0.0249 & 0.1553 \\
    \hdashline
    w$\backslash$o ID fusing & 23.559 & 0.8065 & 0.0183 & 0.0721 \\
    w$\backslash$o Projector & 21.840 & 0.7447 & 0.0216 & 0.1258 \\
    w$\backslash$o LoRA  & 18.953 & 0.7406 & 0.0262 & 0.1570 \\
    \bottomrule
    \end{tabular}
    }
    \label{tab:ablative}
\end{table}

Quantitative comparisons demonstrate that NumeriKontrol achieves superior performance on both low-level and high-level tasks, validating the generalization capability of our method in unseen scenarios. LLM-based methods, including Seedream and Nano Banana, exhibit stronger performance on high-level tasks compared to low-level tasks, which stems from their semantic understanding of input images and instruction comprehension. Conversely, FLUX.1 Kontext achieves higher scores on low-level tasks, which we attribute to Kontext's robust prior for preserving overall image structure. Kontinuous fails to follow precise instructions, but keeps a coarse track on number sequences.

\subsection{Ablative Study}

\begin{figure}[t]
    \centering
    \includegraphics[width=\linewidth]{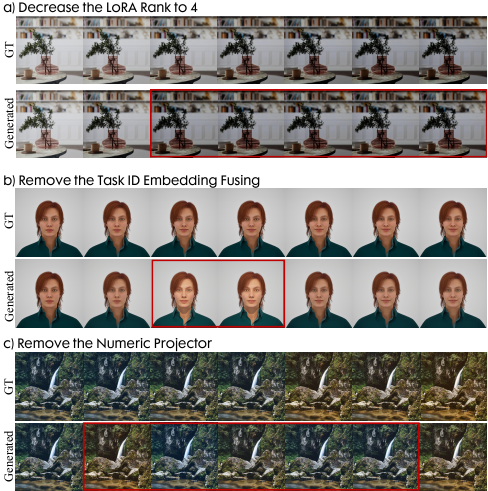}
    \captionof{figure}{\textbf{Visualization of ablative study.} \textbf{a)} Gamma adjustment from 1.0 to 3.0. \textbf{b)} Without task embedding, cross-attribute interference occurs due to overlapping value ranges. \textbf{c)} Adjust the color hue. The relationship between generated images and ground truth(GT) is disordered without Numeric Projector.
    }
    \label{fig:ablative}
\end{figure}

To comprehensively evaluate the importance of each module in NumeriKontrol, we conduct ablation studies on our method. Quantitative results are presented in Table.~\ref{tab:ablative}. We first preserve the Numeric Projector and progressively reduce the rank of LoRA introduced in Sec.~\ref{subsec:conditioning}. As the rank decreases, the trained model gradually loses the correspondence between numeric values and actual visual effects, as shown in Fig.~\ref{fig:ablative}~(a). When LoRA is completely removed, the model's performance nearly degrades to that of standard FLUX.1 Kontext. When the rank decreases to 4, the model converges to only 1-2 cases per instruction template, demonstrating insufficient capacity for numeric control.

For the next ablation, we  preserve the LoRA module and remove the task ID embedding from the Numeric Projector. This operation significantly impacts the MSE metric. As illustrated in Fig.~\ref{fig:ablative}(b), the generated images are influenced by unrelated attribute edits present in the dataset. We further remove the Numeric Projector entirely, retaining only the LoRA modules. In this configuration, the model learns limited representations of numeric information and fails to associate instructions with the desired visual effects for specific attribute values. Visual results are shown in Fig.~\ref{fig:ablative}~(c).

\subsection{Failure Cases}

\begin{figure}[t]
    \centering
    \includegraphics[width=\linewidth]{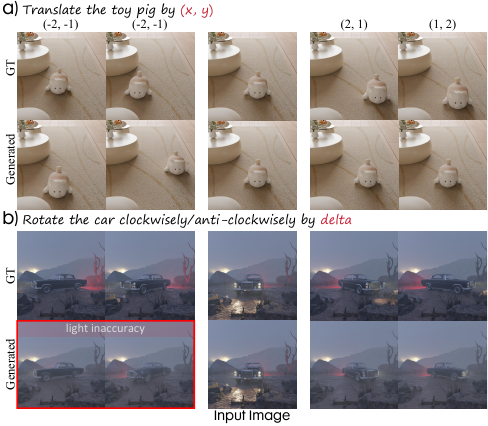}
    \captionof{figure}{\textbf{Visualization of Failure cases.} \textbf{a)} Multi-numeric editing fails to translate an object by the specified vector. \textbf{b)} NumeriKontrol fails to preserve the lighting of the car.
    }
    \label{fig:failure cases}
\end{figure}

NumeriKontrol exhibits limitations in certain tasks. For translation operations, we encode the the displacement $e_x$, $e_y$, $e_z$ into tokens $z_e \in \mathbb{R}^{3 \times 4096}$ and concatenate then with the instruction \textit{Translate the object by ($e_x$, $e_y$, $e_z$)}. We apply a similar approach to Euler angle rotations. Our method fails to establish a clear correspondence between the specified movements and the resulting images as demonstrated in Fig.~\ref{fig:failure cases}~(a). Additional cases occur in scenarios with complex lighting conditions, as shown in Fig.~\ref{fig:failure cases}~(b).

\section{Conclusion}

We present NumeriKontrol, a method that introduces precise numeric control into instruction-based editing. This work makes four key contributions. \textbf{1)} To our knowledge, we are the first to enable instruction-based image editing with explicit numeric parameters. \textbf{2)} We design a lightweight Numeric Adapter to effectively encode parameters with different units. The numeric token integrates into the diffusion model in a plug-and-play manner. \textbf{3)}  We propose a multi-numeric editing paradigm that handles combined instructions. \textbf{4)} We synthesize a large Common Attribute Transformation dataset with physically-accurate content. Each sample is annotated with edited attribute types and scales. This dataset serves as a training resource for NumeriKontrol and benefits the broader research community on similar tasks. Comprehensive experiments demonstrate that NumeriKontrol outperforms state-of-the-art models on numeric-specified editing tasks. Future research will broaden the applicability to more complex editing tasks.
{
    \small
    \bibliographystyle{ieeenat_fullname}
    \bibliography{main}

@String(NIPS= {Adv. Neural Inform. Process. Syst.})

@String(ICLR = {Int. Conf. Learn. Represent.})

@String(AAAI = {AAAI})

@String(NIPS  = {NeurIPS})

@String(ICLR  = {ICLR})

@inproceedings{DDPM,
author = {Ho, Jonathan and Jain, Ajay and Abbeel, Pieter},
title = {Denoising diffusion probabilistic models},
year = {2020},
isbn = {9781713829546},
publisher = {Curran Associates Inc.},
address = {Red Hook, NY, USA},
booktitle = {Proceedings of the 34th International Conference on Neural Information Processing Systems},
articleno = {574},
numpages = {12},
location = {Vancouver, BC, Canada},
series = {NIPS '20}
}

@inproceedings{SD,
  title={High-resolution image synthesis with latent diffusion models},
  author={Rombach, Robin and Blattmann, Andreas and Lorenz, Dominik and Esser, Patrick and Ommer, Bj{\"o}rn},
  booktitle={Proceedings of the IEEE/CVF conference on computer vision and pattern recognition},
  pages={10684--10695},
  year={2022}
}

@inproceedings{IP2P,
  title={Instructpix2pix: Learning to follow image editing instructions},
  author={Brooks, Tim and Holynski, Aleksander and Efros, Alexei A},
  booktitle={Proceedings of the IEEE/CVF conference on computer vision and pattern recognition},
  pages={18392--18402},
  year={2023}
}

@article{styleflow,
  title={Styleflow: Attribute-conditioned exploration of stylegan-generated images using conditional continuous normalizing flows},
  author={Abdal, Rameen and Zhu, Peihao and Mitra, Niloy J and Wonka, Peter},
  journal={ACM Transactions on Graphics},
  volume={40},
  number={3},
  pages={1--21},
  year={2021},
  publisher={ACM New York, NY}
}

@inproceedings{ganspace,
author = {H\"{a}rk\"{o}nen, Erik and Hertzmann, Aaron and Lehtinen, Jaakko and Paris, Sylvain},
title = {GANSpace: discovering interpretable GAN controls},
year = {2020},
isbn = {9781713829546},
publisher = {Curran Associates Inc.},
address = {Red Hook, NY, USA},
booktitle = {Proceedings of the 34th International Conference on Neural Information Processing Systems},
articleno = {825},
numpages = {10},
location = {Vancouver, BC, Canada},
series = {NIPS '20}
}

@inproceedings{styleclip,
  title={Styleclip: Text-driven manipulation of stylegan imagery},
  author={Patashnik, Or and Wu, Zongze and Shechtman, Eli and Cohen-Or, Daniel and Lischinski, Dani},
  booktitle={Proceedings of the IEEE/CVF international conference on computer vision},
  pages={2085--2094},
  year={2021}
}

@inproceedings{marble,
  title={MARBLE: Material Recomposition and Blending in CLIP-Space},
  author={Cheng, Ta Ying and Sharma, Prafull and Boss, Mark and Jampani, Varun},
  booktitle={Proceedings of the Computer Vision and Pattern Recognition Conference},
  pages={13061--13071},
  year={2025}
}

@inproceedings{precisecontrol,
  title={Precisecontrol: Enhancing text-to-image diffusion models with fine-grained attribute control},
  author={Parihar, Rishubh and Sachidanand, VS and Mani, Sabariswaran and Karmali, Tejan and Venkatesh Babu, R},
  booktitle={European Conference on Computer Vision},
  pages={469--487},
  year={2024},
  organization={Springer}
}

@inproceedings{lora,
  title={Lora: Low-rank adaptation of large language models.},
  author={Hu, Edward J and Shen, Yelong and Wallis, Phillip and Allen-Zhu, Zeyuan and Li, Yuanzhi and Wang, Shean and Wang, Lu and Chen, Weizhu and others},
  booktitle={The Tenth International Conference on Learning Representations},
  volume={1},
  number={2},
  pages={3},
  year={2022}
}

@inproceedings{stylegan24,
  title={When stylegan meets stable diffusion: a w+ adapter for personalized image generation},
  author={Li, Xiaoming and Hou, Xinyu and Loy, Chen Change},
  booktitle={Proceedings of the IEEE/CVF Conference on Computer Vision and Pattern Recognition},
  pages={2187--2196},
  year={2024}
}

@inproceedings{alchemist,
  title={Alchemist: Parametric control of material properties with diffusion models},
  author={Sharma, Prafull and Jampani, Varun and Li, Yuanzhen and Jia, Xuhui and Lagun, Dmitry and Durand, Fredo and Freeman, Bill and Matthews, Mark},
  booktitle={Proceedings of the IEEE/CVF Conference on Computer Vision and Pattern Recognition},
  pages={24130--24141},
  year={2024}
}

@article{kontinuous,
  title={Kontinuous Kontext: Continuous Strength Control for Instruction-based Image Editing},
  author={Parihar, Rishubh and Patashnik, Or and Ostashev, Daniil and Babu, R Venkatesh and Cohen-Or, Daniel and Wang, Kuan-Chieh},
  journal={arXiv preprint arXiv:2510.08532},
  year={2025}
}

@article{kontext,
  title={FLUX. 1 Kontext: Flow Matching for In-Context Image Generation and Editing in Latent Space},
  author={Labs, Black Forest and Batifol, Stephen and Blattmann, Andreas and Boesel, Frederic and Consul, Saksham and Diagne, Cyril and Dockhorn, Tim and English, Jack and English, Zion and Esser, Patrick and others},
  journal={arXiv preprint arXiv:2506.15742},
  year={2025}
}

@article{seedream,
  title={Seedream 4.0: Toward next-generation multimodal image generation},
  author={Seedream, Team and Chen, Yunpeng and Gao, Yu and Gong, Lixue and Guo, Meng and Guo, Qiushan and Guo, Zhiyao and Hou, Xiaoxia and Huang, Weilin and Huang, Yixuan and others},
  journal={arXiv preprint arXiv:2509.20427},
  year={2025}
}

@inproceedings{ddim,
  author       = {Jiaming Song and
                  Chenlin Meng and
                  Stefano Ermon},
  title        = {Denoising Diffusion Implicit Models},
  booktitle    = {9th International Conference on Learning Representations, {ICLR} 2021,
                  Virtual Event, Austria, May 3-7, 2021},
  publisher    = {OpenReview.net},
  year         = {2021},
  url          = {https://openreview.net/forum?id=St1giarCHLP},
  bibsource    = {dblp computer science bibliography, https://dblp.org}
}

@inproceedings{photorealistic,
author = {Saharia, Chitwan and Chan, William and Saxena, Saurabh and Lit, Lala and Whang, Jay and Denton, Emily and Ghasemipour, Seyed Kamyar Seyed and Ayan, Burcu Karagol and Mahdavi, S. Sara and Gontijo-Lopes, Raphael and Salimans, Tim and Ho, Jonathan and Fleet, David J and Norouzi, Mohammad},
title = {Photorealistic text-to-image diffusion models with deep language understanding},
year = {2022},
isbn = {9781713871088},
publisher = {Curran Associates Inc.},
address = {Red Hook, NY, USA},
booktitle = {Proceedings of the 36th International Conference on Neural Information Processing Systems},
articleno = {2643},
numpages = {16},
location = {New Orleans, LA, USA},
series = {NIPS '22}
}

@article{sdxl,
  title={Sdxl: Improving latent diffusion models for high-resolution image synthesis},
  author={Podell, Dustin and English, Zion and Lacey, Kyle and Blattmann, Andreas and Dockhorn, Tim and M{\"u}ller, Jonas and Penna, Joe and Rombach, Robin},
  journal={arXiv preprint arXiv:2307.01952},
  year={2023}
}

@inproceedings{masactrl,
  title={Masactrl: Tuning-free mutual self-attention control for consistent image synthesis and editing},
  author={Cao, Mingdeng and Wang, Xintao and Qi, Zhongang and Shan, Ying and Qie, Xiaohu and Zheng, Yinqiang},
  booktitle={Proceedings of the IEEE/CVF international conference on computer vision},
  pages={22560--22570},
  year={2023}
}

@inproceedings{imagic,
  title={Imagic: Text-based real image editing with diffusion models},
  author={Kawar, Bahjat and Zada, Shiran and Lang, Oran and Tov, Omer and Chang, Huiwen and Dekel, Tali and Mosseri, Inbar and Irani, Michal},
  booktitle={Proceedings of the IEEE/CVF conference on computer vision and pattern recognition},
  pages={6007--6017},
  year={2023}
}

@inproceedings{clip,
  title={Learning transferable visual models from natural language supervision},
  author={Radford, Alec and Kim, Jong Wook and Hallacy, Chris and Ramesh, Aditya and Goh, Gabriel and Agarwal, Sandhini and Sastry, Girish and Askell, Amanda and Mishkin, Pamela and Clark, Jack and others},
  booktitle={Proceedings of the 38th International Conference on Machine Learning},
  pages={8748--8763},
  year={2021},
  organization={PmLR}
}

@inproceedings{rf,
  title={Scaling rectified flow transformers for high-resolution image synthesis},
  author={Esser, Patrick and Kulal, Sumith and Blattmann, Andreas and Entezari, Rahim and M{\"u}ller, Jonas and Saini, Harry and Levi, Yam and Lorenz, Dominik and Sauer, Axel and Boesel, Frederic and others},
  booktitle={Forty-first international conference on machine learning},
  year={2024}
}

@misc{flux2024,
    author={Black Forest Labs},
    title={FLUX},
    year={2024},
    howpublished={\url{https://github.com/black-forest-labs/flux}},
}

@inproceedings{dit,
  title={Scalable diffusion models with transformers},
  author={Peebles, William and Xie, Saining},
  booktitle={Proceedings of the IEEE/CVF international conference on computer vision},
  pages={4195--4205},
  year={2023}
}

@inproceedings{controlnet,
  title={Adding conditional control to text-to-image diffusion models},
  author={Zhang, Lvmin and Rao, Anyi and Agrawala, Maneesh},
  booktitle={Proceedings of the IEEE/CVF international conference on computer vision},
  pages={3836--3847},
  year={2023}
}

@inproceedings{t2i-adapter,
author = {Mou, Chong and Wang, Xintao and Xie, Liangbin and Wu, Yanze and Zhang, Jian and Qi, Zhongang and Shan, Ying},
title = {T2I-Adapter: learning adapters to dig out more controllable ability for text-to-image diffusion models},
year = {2024},
isbn = {978-1-57735-887-9},
publisher = {AAAI Press},
url = {https://doi.org/10.1609/aaai.v38i5.28226},
doi = {10.1609/aaai.v38i5.28226},
booktitle = {Proceedings of the Thirty-Eighth AAAI Conference on Artificial Intelligence and Thirty-Sixth Conference on Innovative Applications of Artificial Intelligence and Fourteenth Symposium on Educational Advances in Artificial Intelligence},
articleno = {478},
numpages = {9},
series = {AAAI'24/IAAI'24/EAAI'24}
}

@inproceedings{easycontrol,
  title={Easycontrol: Adding efficient and flexible control for diffusion transformer},
  author={Zhang, Yuxuan and Yuan, Yirui and Song, Yiren and Wang, Haofan and Liu, Jiaming},
  booktitle={Proceedings of the IEEE/CVF International Conference on Computer Vision},
  pages={19513--19524},
  year={2025}
}

@inproceedings{ominicontrol,
  title={Ominicontrol: Minimal and universal control for diffusion transformer},
  author={Tan, Zhenxiong and Liu, Songhua and Yang, Xingyi and Xue, Qiaochu and Wang, Xinchao},
  booktitle={Proceedings of the IEEE/CVF International Conference on Computer Vision},
  pages={14940--14950},
  year={2025}
}

@article{wang2025diffdecompose,
  title={DiffDecompose: Layer-Wise Decomposition of Alpha-Composited Images via Diffusion Transformers},
  author={Wang, Zitong and Zhao, Hang and Zhou, Qianyu and Lu, Xuequan and Li, Xiangtai and Song, Yiren},
  journal={arXiv preprint arXiv:2505.21541},
  year={2025}
}

@article{gong2025relationadapter,
  title={RelationAdapter: Learning and Transferring Visual Relation with Diffusion Transformers},
  author={Gong, Yan and Song, Yiren and Li, Yicheng and Li, Chenglin and Zhang, Yin},
  journal={arXiv preprint arXiv:2506.02528},
  year={2025}
}

@article{jiang2025personalized,
  title={Personalized Vision via Visual In-Context Learning},
  author={Jiang, Yuxin and Gu, Yuchao and Song, Yiren and Tsang, Ivor and Shou, Mike Zheng},
  journal={arXiv preprint arXiv:2509.25172},
  year={2025}
}

@article{song2025layertracer,
  title={LayerTracer: Cognitive-Aligned Layered SVG Synthesis via Diffusion Transformer},
  author={Song, Yiren and Chen, Danze and Shou, Mike Zheng},
  journal={arXiv preprint arXiv:2502.01105},
  year={2025}
}

@article{song2025makeanything,
  title={MakeAnything: Harnessing Diffusion Transformers for Multi-Domain Procedural Sequence Generation},
  author={Song, Yiren and Liu, Cheng and Shou, Mike Zheng},
  journal={arXiv preprint arXiv:2502.01572},
  year={2025}
}

@article{huang2025photodoodle,
  title={Photodoodle: Learning artistic image editing from few-shot pairwise data},
  author={Huang, Shijie and Song, Yiren and Zhang, Yuxuan and Guo, Hailong and Wang, Xueyin and Shou, Mike Zheng and Liu, Jiaming},
  journal={arXiv preprint arXiv:2502.14397},
  year={2025}
}

@article{song2025omniconsistency,
  title={Omniconsistency: Learning style-agnostic consistency from paired stylization data},
  author={Song, Yiren and Liu, Cheng and Shou, Mike Zheng},
  journal={arXiv preprint arXiv:2505.18445},
  year={2025}
}

@article{guo2025any2anytryon,
  title={Any2AnyTryon: Leveraging Adaptive Position Embeddings for Versatile Virtual Clothing Tasks},
  author={Guo, Hailong and Zeng, Bohan and Song, Yiren and Zhang, Wentao and Zhang, Chuang and Liu, Jiaming},
  journal={arXiv preprint arXiv:2501.15891},
  year={2025}
}

@article{chen2025transanimate,
  title={Transanimate: Taming layer diffusion to generate rgba video},
  author={Chen, Xuewei and Chen, Zhimin and Song, Yiren},
  journal={arXiv preprint arXiv:2503.17934},
  year={2025}
}

@article{song2024processpainter,
  title={Processpainter: Learn painting process from sequence data},
  author={Song, Yiren and Huang, Shijie and Yao, Chen and Ye, Xiaojun and Ci, Hai and Liu, Jiaming and Zhang, Yuxuan and Shou, Mike Zheng},
  journal={arXiv preprint arXiv:2406.06062},
  year={2024}
}

@inproceedings{zhang2025stable,
  title={Stable-hair: Real-world hair transfer via diffusion model},
  author={Zhang, Yuxuan and Zhang, Qing and Song, Yiren and Zhang, Jichao and Tang, Hao and Liu, Jiaming},
  booktitle={Proceedings of the AAAI Conference on Artificial Intelligence},
  volume={39},
  number={10},
  pages={10348--10356},
  year={2025}
}

@article{zhang2024stable,
  title={Stable-makeup: When real-world makeup transfer meets diffusion model},
  author={Zhang, Yuxuan and Wei, Lifu and Zhang, Qing and Song, Yiren and Liu, Jiaming and Li, Huaxia and Tang, Xu and Hu, Yao and Zhao, Haibo},
  journal={arXiv preprint arXiv:2403.07764},
  year={2024}
}

@article{ipadapter,
  title={Ip-adapter: Text compatible image prompt adapter for text-to-image diffusion models},
  author={Ye, Hu and Zhang, Jun and Liu, Sibo and Han, Xiao and Yang, Wei},
  journal={arXiv preprint arXiv:2308.06721},
  year={2023}
}

@article{prompt2prompt,
  title={Prompt-to-prompt image editing with cross attention control},
  author={Hertz, Amir and Mokady, Ron and Tenenbaum, Jay and Aberman, Kfir and Pritch, Yael and Cohen-Or, Daniel},
  journal={arXiv preprint arXiv:2208.01626},
  year={2022}
}

@article{qwen-image,
  title={Qwen-image technical report},
  author={Wu, Chenfei and Li, Jiahao and Zhou, Jingren and Lin, Junyang and Gao, Kaiyuan and Yan, Kun and Yin, Sheng-ming and Bai, Shuai and Xu, Xiao and Chen, Yilei and others},
  journal={arXiv preprint arXiv:2508.02324},
  year={2025}
}

@article{omnigen2,
  title={OmniGen2: Exploration to Advanced Multimodal Generation},
  author={Wu, Chenyuan and Zheng, Pengfei and Yan, Ruiran and Xiao, Shitao and Luo, Xin and Wang, Yueze and Li, Wanli and Jiang, Xiyan and Liu, Yexin and Zhou, Junjie and others},
  journal={arXiv preprint arXiv:2506.18871},
  year={2025}
}

@article{ma2025followfaster,
  title={Follow-your-emoji-faster: Towards efficient, fine-controllable, and expressive freestyle portrait animation},
  author={Ma, Yue and Yan, Zexuan and Liu, Hongyu and Wang, Hongfa and Pan, Heng and He, Yingqing and Yuan, Junkun and Zeng, Ailing and Cai, Chengfei and Shum, Heung-Yeung and others},
  journal={arXiv preprint arXiv:2509.16630},
  year={2025}
}

@article{ma2025controllable,
  title={Controllable Video Generation: A Survey},
  author={Ma, Yue and Feng, Kunyu and Hu, Zhongyuan and Wang, Xinyu and Wang, Yucheng and Zheng, Mingzhe and He, Xuanhua and Zhu, Chenyang and Liu, Hongyu and He, Yingqing and others},
  journal={arXiv preprint arXiv:2507.16869},
  year={2025}
}

@article{ma2023magicstick,
  title={Magicstick: Controllable video editing via control handle transformations},
  author={Ma, Yue and Cun, Xiaodong and He, Yingqing and Qi, Chenyang and Wang, Xintao and Shan, Ying and Li, Xiu and Chen, Qifeng},
  journal={arXiv preprint arXiv:2312.03047},
  year={2023}
}

@inproceedings{ma2024followpose,
  title={Follow your pose: Pose-guided text-to-video generation using pose-free videos},
  author={Ma, Yue and He, Yingqing and Cun, Xiaodong and Wang, Xintao and Chen, Siran and Li, Xiu and Chen, Qifeng},
  booktitle={Proceedings of the AAAI Conference on Artificial Intelligence},
  volume={38},
  number={5},
  pages={4117--4125},
  year={2024}
}

@inproceedings{ma2022visual,
  title={Visual knowledge graph for human action reasoning in videos},
  author={Ma, Yue and Wang, Yali and Wu, Yue and Lyu, Ziyu and Chen, Siran and Li, Xiu and Qiao, Yu},
  booktitle={Proceedings of the 30th ACM International Conference on Multimedia},
  pages={4132--4141},
  year={2022}
}

@article{ma2025followcreation,
  title={Follow-Your-Creation: Empowering 4D Creation through Video Inpainting},
  author={Ma, Yue and Feng, Kunyu and Zhang, Xinhua and Liu, Hongyu and Zhang, David Junhao and Xing, Jinbo and Zhang, Yinhan and Yang, Ayden and Wang, Zeyu and Chen, Qifeng},
  journal={arXiv preprint arXiv:2506.04590},
  year={2025}
}

@article{ma2025followyourmotion,
  title={Follow-Your-Motion: Video Motion Transfer via Efficient Spatial-Temporal Decoupled Finetuning},
  author={Ma, Yue and Liu, Yulong and Zhu, Qiyuan and Yang, Ayden and Feng, Kunyu and Zhang, Xinhua and Li, Zhifeng and Han, Sirui and Qi, Chenyang and Chen, Qifeng},
  journal={arXiv preprint arXiv:2506.05207},
  year={2025}
}

@inproceedings{ma2024followyouremoji,
  title={Follow-your-emoji: Fine-controllable and expressive freestyle portrait animation},
  author={Ma, Yue and Liu, Hongyu and Wang, Hongfa and Pan, Heng and He, Yingqing and Yuan, Junkun and Zeng, Ailing and Cai, Chengfei and Shum, Heung-Yeung and Liu, Wei and others},
  booktitle={SIGGRAPH Asia 2024 Conference Papers},
  pages={1--12},
  year={2024}
}

@inproceedings{ma2025followyourclick,
  title={Follow-Your-Click: Open-domain Regional Image Animation via Motion Prompts},
  author={Ma, Yue and He, Yingqing and Wang, Hongfa and Wang, Andong and Shen, Leqi and Qi, Chenyang and Ying, Jixuan and Cai, Chengfei and Li, Zhifeng and Shum, Heung-Yeung and others},
  booktitle={Proceedings of the AAAI Conference on Artificial Intelligence},
  volume={39},
  number={6},
  pages={6018--6026},
  year={2025}
}

@article{lu2025easytext,
  title={EasyText: Controllable Diffusion Transformer for Multilingual Text Rendering},
  author={Lu, Runnan and Zhang, Yuxuan and Liu, Jiaming and Wang, Haofan and Song, Yiren},
  journal={arXiv preprint arXiv:2505.24417},
  year={2025}
}

@article{shi2024fonts,
  title={FonTS: Text Rendering with Typography and Style Controls},
  author={Shi, Wenda and Song, Yiren and Zhang, Dengming and Liu, Jiaming and Zou, Xingxing},
  journal={arXiv preprint arXiv:2412.00136},
  year={2024}
}

@article{shi2025wordcon,
  title={WordCon: Word-level Typography Control in Scene Text Rendering},
  author={Shi, Wenda and Song, Yiren and Rao, Zihan and Zhang, Dengming and Liu, Jiaming and Zou, Xingxing},
  journal={arXiv preprint arXiv:2506.21276},
  year={2025}
}

@article{cfg,
  title={Classifier-free diffusion guidance},
  author={Ho, Jonathan and Salimans, Tim},
  journal={arXiv preprint arXiv:2207.12598},
  year={2022}
}

@ARTICLE{ssim,
  author={Zhou Wang and Bovik, A.C. and Sheikh, H.R. and Simoncelli, E.P.},
  journal={IEEE Transactions on Image Processing}, 
  title={Image quality assessment: from error visibility to structural similarity}, 
  year={2004},
  volume={13},
  number={4},
  pages={600-612},
  doi={10.1109/TIP.2003.819861}
}

@inproceedings{lpips,
  title={The unreasonable effectiveness of deep features as a perceptual metric},
  author={Zhang, Richard and Isola, Phillip and Efros, Alexei A and Shechtman, Eli and Wang, Oliver},
  booktitle={Proceedings of the IEEE conference on computer vision and pattern recognition},
  pages={586--595},
  year={2018}
}

@inproceedings{diffmorpher,
  title={Diffmorpher: Unleashing the capability of diffusion models for image morphing},
  author={Zhang, Kaiwen and Zhou, Yifan and Xu, Xudong and Dai, Bo and Pan, Xingang},
  booktitle={Proceedings of the IEEE/CVF Conference on Computer Vision and Pattern Recognition},
  pages={7912--7921},
  year={2024}
}

@article{freemorph,
  title={FreeMorph: Tuning-Free Generalized Image Morphing with Diffusion Model},
  author={Cao, Yukang and Si, Chenyang and Wang, Jinghao and Liu, Ziwei},
  journal={arXiv preprint arXiv:2507.01953},
  year={2025}
}
}

\clearpage
\setcounter{page}{1}
\maketitlesupplementary

\appendix

\section{User Study}
\label{sec:user_study}

To further demonstrate the effectiveness of NumeriKontrol, we conduct a user study comparing five methods: NumeriKontrol, Kontinuous~\cite{kontinuous}, FLUX.1 Kontext~\cite{kontext}, Nano Banana, and Seedream 4.0~\cite{seedream}. We sample 30 examples for each method. As described in Sec.~\ref{subsec:qualitative}, Nano Banana and Seedream do not support specific numerical instructions. All instructions are adjusted and tested to ensure consistent evaluation across methods.

Following standardized evaluation protocols, we anonymize all samples by removing method identifiers and randomize their presentation order. The study is conducted through an online questionnaire distributed to participants. Participants view examples from all five methods and rate each sample on a 1-5 Likert scale (allowing one decimal place). The evaluation includes three criteria: (1) \textit{Is the image edited according to the instruction?} (2) \textit{Does the edited result match the numeric information in the instruction?} (3) \textit{How much do you prefer the edited result?} These questions correspond to success rate, numerical alignment, and user preference, respectively.

\begin{table}[h]
    \centering
    \caption{User study with state-of-the-art methods. The best score is \textbf{emphasized}.}
    \resizebox{\linewidth}{!}{
    \begin{tabular}{lccc}
        \toprule 
        Method &  \thead{Success} ($\uparrow$) & \thead{Alignment} ($\uparrow$) & \thead{Preference} ($\uparrow$)\\ 
        \hline  \hline 

        FLUX.1 Kontext~\cite{kontext} & 3.64   & 2.70 &  3.57 \\
        \hline 
        
        Kontinuous Kontext~\cite{kontinuous} &  4.75 & 4.24 & 3.85 \\
        \hline   

        Nano Banana & 4.42 & 3.25 & 3.62 \\
        \hline   

        Seedream~\cite{seedream} & 4.31 & 3.09 & 3.78 \\
        \hline   
        
        NumeriKontrol(Ours)  & \textbf{4.80} & \textbf{4.57} & \textbf{4.08} \\
        \bottomrule
        
    \end{tabular}
    }
    \label{table:user-study}
\end{table}

The results of the user study is shown in Tab.~\ref{table:user-study}. Our method scores best in all three criteria.

\section{More Details for Comparison}

\subsection{Implementation Details for Comparison}

As Kontinuous Kontext~\cite{kontinuous} is implemented by ourself, the strength projector in our implementation is designed with dimensions of 1536 $\rightarrow$ 256 $\rightarrow$ 128 $\rightarrow$ 6144. The scale and shift are separated into 3072 dimensions respectively from the output of the projector. The modulation parameter is set to 1.0 in our experiments.

\subsection{Quantitative Comparison}

Certain prompts may not function properly with Nano Banana and Seedream. To address this limitation, we develop adapted instruction formats for these methods. For camera-related aspects in the test set, we append \textit{"Work as a DSLR camera"} to the instructions to enable editing. For other numeric instructions, we adopt the following format: \textit{"Edit the image with a strength of $\delta$. The full strength is $\Delta$}, where $\delta$ represents the current editing strength and $\Delta$ denotes the maximum strength. Details of the numeric instructions are provided in Sec.~\ref{subsec:dataset}. All instructions used in quantitative comparisons are standardized using these two approaches to ensure fair evaluation.

\begin{figure}[t]
    \centering
    \includegraphics[width=\linewidth]{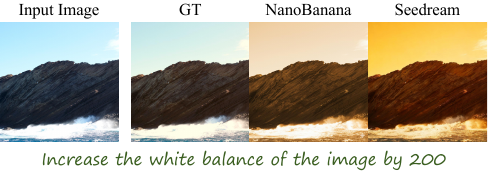}
    \captionof{figure}{The low-level task of editing the white balance of the image. The white balance is divided into 400 steps, with 200 lower and 200 higher.
    }
    \label{fig:quantitative-example}
\end{figure}

As described in Sec.\ref{subsec:quantitative}, LLM-based methods successfully edit images for low-level tasks. However, they fail to preserve accurate numeric information from the instructions. Consequently, these two methods perform poorly on low-level tasks. An example from the quantitative results is shown in Fig.\ref{fig:quantitative-example}.

\begin{figure}[b]
    \centering
    \includegraphics[width=\linewidth]{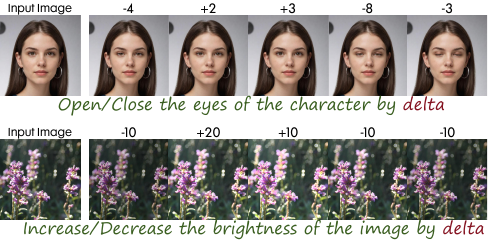}
    \captionof{figure}{\textbf{Visualization of Continuous Editing.}
    }
    \label{fig:continuous}
\end{figure}

\section{Additional Results}

Our method supports continuous editing through the online delta calculation described in Sec.\ref{subsec:conditioning}. However, errors may accumulate during successive operations. As illustrated in Fig.\ref{fig:continuous}, despite the first row's cumulative parameter changes totaling -10, the final state exhibits incomplete closure; conversely, the second row achieves zero net change yet displays reduced brightness. While continuous editing cannot guarantee absolute numerical precision, the overall directional trends remain consistent.

\section{More Details on Synthesized Dataset}

\begin{figure*}[t]
    \centering
    \includegraphics[width=\linewidth]{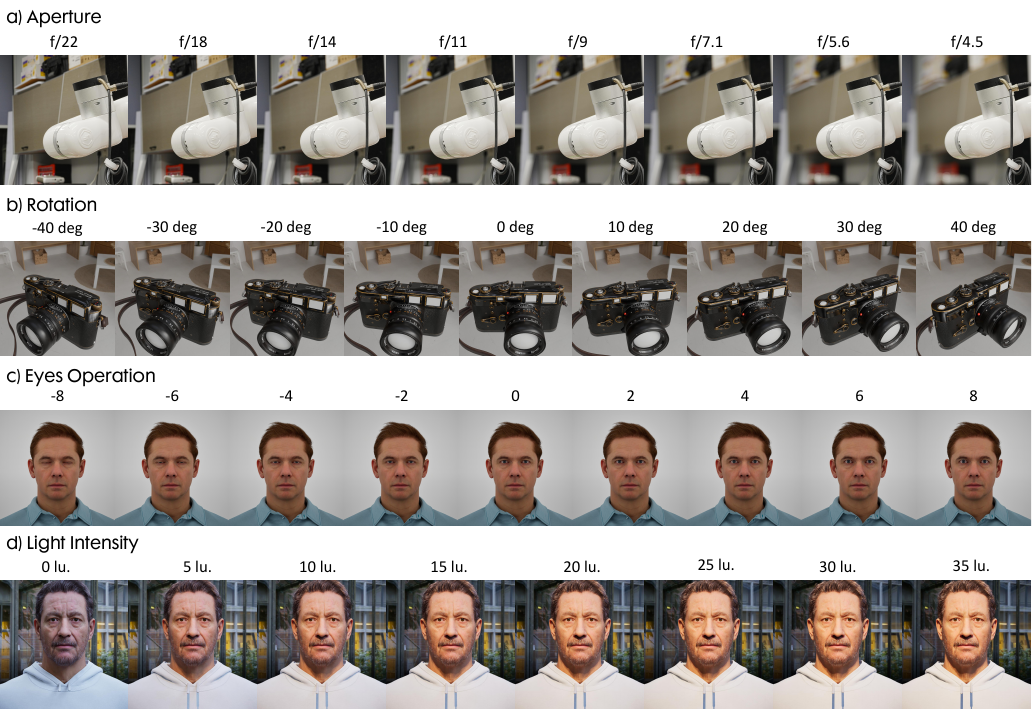}
    \captionof{figure}{
    \textbf{Visualization of samples in CAT dataset.}
    }
    \label{fig:dataset_samples}
\end{figure*}

We illustrate samples from our Common Attribute Transformation dataset in Fig.\ref{fig:dataset_samples}. All samples are annotated with accurate attribute values. For facial expression tasks such as eye opening (Fig.\ref{fig:dataset_samples}~(c)), we use a simple gray HDRI for training. This choice does not interfere with the model's generalization ability.
However, for lighting-related tasks such as intensity adjustment (Fig.\ref{fig:dataset_samples}(d)), we employ diverse HDRI backgrounds. This design reflects the complexity of real-world lighting conditions. Multiple HDRIs approximate these complex light sources and improve generalization performance.
For object manipulation scenarios (Fig.\ref{fig:dataset_samples}(b)), the object name is inserted into the instruction template. All low-level tasks follow a unified prompt format described in Sec.\ref{subsec:dataset}, as shown in Fig.\ref{fig:dataset_samples}~(a). All images in our dataset is taken in 1024$\times$1024. During training, the samples are resized to 512$\times$512.

\end{document}